\PassOptionsToPackage{expansion=false}{microtype}
\documentclass[10pt, a4paper]{article}
\usepackage[final]{lrec2026}

\usepackage[T1]{fontenc}
\usepackage{CJKutf8}

\newcommand{\langjp}[1]{%
\begin{CJK*}{UTF8}{min}%
\microtypecontext{activate=false}
#1%
\end{CJK*}%
}

\usepackage{graphicx}
\usepackage{booktabs}
\usepackage{tabularx}
\usepackage{multirow}
\usepackage{amsmath}
\usepackage{hyperref}
\usepackage[table]{xcolor}

\DeclareUnicodeCharacter{202F}{\ }

\title{Mind Your Moras: Orthography-Aware Error Analysis of \\ Neural Japanese Morphological Generation}
\name{Wen Zhang}
\address{\href{mailto:wenzhang0222@gmail.com}{\texttt{wenzhang0222@gmail.com}}}

\begin{document}

\begin{abstract}
{
We present an orthography-aware error analysis of Japanese past-tense morphological inflection, treating \textit{hiragana} not merely as a transcriptional medium, but as a representational system encoding morphophonological distinctions that may influence model generalization. We evaluate two character-level sequence-to-sequence architectures on past-tense formation using datasets formatted according to the SIGMORPHON 2020 and 2023 shared task conventions. Despite high aggregate accuracy, models exhibit systematic, linguistically interpretable errors that cluster around specific orthographic properties of \textit{hiragana}. We introduce a concise error taxonomy capturing seven primary failure modes and provide both quantitative and qualitative analyses. Gemination-related errors dominate residual failures, accounting for 75--80\% of errors, particularly in verbs whose stems end in /e/ and require gemination before the past-tense suffix. Error patterns remain highly consistent across architectures and random seeds, suggesting a robust interaction between orthographic representation, morphological structure, and data frequency effects in shaping model generalization. These results underscore the necessity of orthography-aware evaluation for understanding neural generalization in morphologically complex languages.
\\
\newline
\Keywords{morphological inflection, error analysis, Japanese \textit{hiragana}, orthographic representation, character-level models, morphophonology}
}
\end{abstract}

\maketitleabstract

\section{Introduction}

Japanese verbs are written in a hybrid system of \textit{kanji}, \textit{hiragana}, and \textit{katakana}. While verb stems may appear in either \textit{kanji} or \textit{hiragana}, inflectional suffixes are consistently marked in \textit{hiragana}. For example, the verb \langjp{書く} / \langjp{かく} \textit{kaku} → \langjp{書いた} / \langjp{かいた} \textit{kaita} in the past tense, where \langjp{た} \textit{ta} marks the past-tense suffix. \textit{Hiragana} thus provides a controlled layer of morphophonological information that can influence model generalization. In this study, we focus on \textit{hiragana} as a computationally relevant layer of linguistic representation, using past-tense inflection as a case study to analyze orthography-sensitive patterns. By restricting all verb forms to \textit{hiragana}, we eliminate confounds from \textit{kanji} homography and lexical ambiguity, creating a controlled environment to investigate how neural models capture orthography-sensitive patterns such as gemination, vowel lengthening, and moraic timing \citep{Vance1989}. Our goal is not to optimize inflection performance per se, but to analyze systematic error patterns associated with orthographic representation.

Most evaluations of Japanese morphological generation focus on aggregate accuracy metrics rather than analyzing how errors relate to orthographic representation. Prior shared tasks primarily report aggregate exact-match accuracy for morphological inflection \citep{CotterellEtAl2016, VylomovaEtAl2020, GoldmanEtAl2023}, often exceeding 95\% for high-resource languages such as Japanese. However, high aggregate accuracy does not necessarily indicate robust generalization across models and training regimes \citep{MakarovClematide2018}. Systematic orthography-sensitive patterns remain underexplored, despite the fact that \textit{hiragana} systematically encodes moraic and morphophonological distinctions central to Japanese phonology. As a moraic script, \textit{hiragana} represents phonological timing units and marks processes such as gemination (the small \langjp{っ} \textit{tsu}) and vowel lengthening, which play a central role in Japanese morphophonology \citep{KubozonoEtAl2008, Labrune2012}. Following the view that writing systems constitute structured representational systems rather than transparent encodings of speech \citep{Sproat2000, DanielsBright1996, Zhang-2023}, we treat \textit{hiragana} as a structured representational system that may shape model generalization.

Japanese past-tense morphology, realized via the suffix -\textit{ta}, varies systematically with phonological and orthographic properties of the verb stem \citep{Vance1989}. These rule-governed and lexically conditioned alternations provide a natural testbed for analyzing orthography-sensitive patterns.

Traditionally, Japanese verbs are classified by inflectional behavior:

\textbf{\textit{Godan} (\textit{u}-) verbs} – form the past tense through suffix-conditioned stem alternations, often involving consonant changes and gemination:  
\langjp{かく} \textit{kaku} ‘to write’ → \langjp{かいた} \textit{kaita},  
\langjp{たつ} \textit{tatsu} ‘to stand’ → \langjp{たった} \textit{tatta},  
\langjp{のむ} \textit{nomu} ‘to drink’ → \langjp{のんだ} \textit{nonda}.  

\textbf{\textit{Ichidan} (\textit{ru}-) verbs} – exhibit stable stems; past tense is formed through direct suffix attachment:  
\langjp{みる} \textit{miru} ‘to see’ → \langjp{みた} \textit{mita},  
\langjp{たべる} \textit{taberu} ‘to eat’ → \langjp{たべた} \textit{tabeta}. 

\textbf{Irregular verbs} – a small closed class including highly frequent items like \langjp{する} \textit{suru} ‘to do’ and \langjp{くる} \textit{kuru} ‘to come’, which display non-canonical stem alternations:  
\langjp{する} \textit{suru} → \langjp{した} \textit{shita},  
\langjp{くる} \textit{kuru} → \langjp{きた} \textit{kita}.  

These verb classes instantiate morphophonological processes, including stem alternation, voicing, and gemination, directly reflected in \textit{hiragana} orthography. Focusing on past-tense forms highlights orthography-sensitive morphophonological alternations densely encoded in \textit{hiragana}.

\section{Data}

We use a Japanese verb inflection dataset formatted following SIGMORPHON conventions \citep{VylomovaEtAl2020, GoldmanEtAl2023}. All forms are converted to \textit{hiragana} only. Each instance consists of three TAB‑separated fields: lemma, target form, and tag (placeholder \_ to test whether the model can learn the mapping without explicit morphosyntactic features).

\vspace{0.5em}
\textbf{Example:} \texttt{\langjp{ねがえる} \quad \langjp{ねがえった} \quad \_}
\vspace{0.5em}

\subsection{Verb Classification}

Verbs are classified based on traditional Japanese conjugation classes, refined to capture orthography‑sensitive variation. In this framework, Japanese verbs are divided into four types: Type 1 (\textit{Godan}), Type 2 (\textit{Ichidan}), Type 3 (canonical irregular), and Type 4 (other irregular forms).

The canonical irregular verbs \langjp{する} \textit{suru} ‘to do’ and \langjp{くる} \textit{kuru} ‘to come’~(Type 3), as well as polysemous lemmas with multiple inflected forms, are excluded to maintain a clear one-to-one lemma–form mapping.

\textbf{Type 1 (\textit{Godan} verbs):} Regular \textit{u}-verbs exhibiting productive, suffix-conditioned stem alternations.  
Example: \langjp{かく} \textit{kaku} 'to write' → \langjp{かいた} \textit{kaita}.  
Count: 2,503

\textbf{Type 2 (\textit{Ichidan} verbs):} Regular \textit{ru}-verbs with stable stems and predictable suffix attachment.  
Example: \langjp{たべる} \textit{taberu} 'to eat' → \langjp{たべた} \textit{tabeta}.  
Count: 1,298

\textbf{Type 4 (Other irregular verbs):} Verbs that deviate from standard \textit{Godan} or \textit{Ichidan} patterns, exhibiting non-standard orthographic behavior in past-tense formation. Subtypes capture finer-grained orthographic variation:

\begin{itemize}
    \item \textbf{Type 4-1:} Stem alternations involving gemination in verbs with stem-final /i/. These verbs resemble Type 2 but do not fully follow regular patterns. Gemination occurs at the boundary between the stem-final /i/ and the past-tense suffix -\textit{ta}.
    Example: \langjp{まじる} \textit{majiru} → \langjp{まじった} \textit{majitta}.  
    Count: 119

    \item \textbf{Type 4-2:} Stem alternations involving gemination in verbs with stem-final /e/. These verbs also resemble Type 2, but gemination occurs at the boundary between the stem-final /e/ and the past-tense suffix -\textit{ta}.  
    Example: \langjp{あきれかえる} \textit{akirekaeru} → \langjp{あきれかえった} \textit{akirekaetta}.
    Count: 37

    \item \textbf{Type 4-3:} Localized deviations. These verbs largely follow Type 1 formation but contain idiosyncratic stem behavior. Due to the one-to-one lemma-to-form constraint in our dataset, only a single instance appears.  
    Example: \langjp{いく} \textit{iku} → \langjp{いった} \textit{itta}.  
    Count: 1
\end{itemize}

Table~\ref{tab:dataset-verbs} summarizes the dataset statistics across verb types.

\begin{table}[t]
\centering
\begin{tabular}{@{}l r@{}}
\toprule
\textbf{Verb Type} & \textbf{Count} \\
\midrule
All verbs & 3,958 \\
Type 1 (Godan) & 2,503 \\
Type 2 (Ichidan) & 1,298 \\
Type 3 (canonical irregular; excluded) & 0 \\
Type 4 (other irregular) & 157 \\
\hspace{1em}Type 4-1 (stem-final /i/ + gemination) & 119 \\
\hspace{1em}Type 4-2 (stem-final /e/ + gemination) & 37 \\
\hspace{1em}Type 4-3 (localized) & 1 \\
\bottomrule
\end{tabular}
\caption{Dataset statistics by verb type.}
\label{tab:dataset-verbs}
\end{table}

\section{Methods}

\subsection{Models}

We evaluate two neural sequence-to-sequence architectures derived from recent SIGMORPHON shared tasks. Both models operate at the character level and generate inflected forms autoregressively, allowing fine-grained modeling of morphophonological alternations.  

The first model corresponds to the official baseline system for Shared Task 0 at SIGMORPHON 2020 \citep{VylomovaEtAl2020}. This baseline uses a character‑level encoder–decoder Transformer model \citep{VaswaniEtAl2017} with standard attention, trained to map lemma + feature representations to inflected forms. Prior work on morphological string transduction has shown that augmenting sequence‑to‑sequence models with mechanisms that facilitate copying behavior, such as pointer‑generator or monotonic attention, can improve performance on low‑frequency and orthographically conditioned forms \citep{MakarovClematide2018, WuCotterell2019}.  

The second model follows the architecture used in the SIGMORPHON–UniMorph 2023 Shared Task, which also employs a Transformer‑based encoder–decoder framework but is designed to improve generalization to unseen lemmas through a lemma‑split training and evaluation regime \citep{GoldmanEtAl2023}. Under both architectures, the model takes a lemma–feature representation as input and predicts the target inflected form character by character.

\subsection{Training and Evaluation}

Both models are trained on the full dataset, and training proceeds using cross-entropy loss with teacher forcing under the default hyperparameter settings of the respective original systems. Evaluation is conducted using exact-match accuracy at the lemma level, consistent with shared task conventions. In addition to aggregate accuracy, we perform detailed error analysis to identify recurrent, orthography-sensitive failure modes, providing both quantitative and qualitative insight into \textit{hiragana} performance.

\subsection{Experimental Controls and Data Splits}

Neural model performance is sensitive to stochastic factors such as random initialization, data shuffling, and optimization dynamics. Prior work has shown that varying random seeds can produce statistically significant differences, even for state-of-the-art sequence tagging systems \citep{ReimersGurevych2017}.  

To account for this variability and reduce selective reporting bias, we trained each model five times using different random seeds. The dataset was split into 80\% training, 10\% development, and 10\% test sets.

\section{Results}

\subsection{Overall Performance}

Both models achieve high test-set accuracy:

\begin{center}
\textbf{SIGMORPHON 2020:} 97.97\%

\textbf{SIGMORPHON 2023:} 97.17\%
\end{center}

Aggregated across five random seeds per model, we analyze 104 total errors (53 from 2020; 51 from 2023).

\subsection{Error Taxonomy}

We classify residual errors according to orthographic and morphophonological processes, defining seven primary failure modes: gemination omission, gemination insertion, phonological substitution, morpheme boundary errors, character recognition errors, compound verb structural errors, and stem alternation overregularization. Table~\ref{tab:error-taxonomy} summarises these categories along with the corresponding orthographic or phonological properties that characterise them.

This taxonomy highlights structured, writing‑system‑sensitive patterns in model failures. For example, failures to insert or spuriously insert the small \textit{\langjp{っ}} reflect challenges with consonant doubling, while stem alternation overregularization captures cases where models fail to apply context‑sensitive consonant deletion before gemination. These structured error types suggest that residual failures in \textit{hiragana} inflection are systematically associated with specific orthographic and morphophonological phenomena rather than being purely random noise.

\begin{table*}[t]
\centering
\renewcommand{\arraystretch}{1.1}
\begin{tabularx}{\textwidth}{l X l}
\toprule
\textbf{Error Type} & \textbf{Description} & \textbf{Orthographic/Phonological Property} \\
\midrule
Gemination omission
  & Failure to insert small \langjp{っ}
  & Consonant doubling \\

Gemination insertion
  & Spurious insertion of \langjp{っ}
  & Consonant doubling \\

Phonological substitution
  & Incorrect consonant or vowel in stem
  & Sound alternation \\

Morpheme boundary error
  & Misaligned suffix attachment
  & Boundary detection \\

Character recognition error
  & Generation of UNK symbol
  & Encoding/representation \\

Compound verb error
  & Structural collapse in compound verbs
  & Compound segmentation \\

\begin{tabular}[t]{@{}l@{}}
Stem alternation error \\
(overregularization)
\end{tabular}
  & Failure to delete a stem-final consonant before gemination
  & Consonant deletion–gemination interaction \\

\bottomrule
\end{tabularx}
\caption{Error taxonomy for \textit{hiragana}-based inflection, showing primary failure modes and their associated orthographic or phonological properties.}
\label{tab:error-taxonomy}
\end{table*}

\subsection{Quantitative Error Distribution}

Not all conceptual error types were observed with sufficient frequency for quantitative analysis. Table~\ref{tab:error-quant} reports the six most frequently occurring error types, aggregated across five random seeds. Gemination-related errors dominate, accounting for 75–80\% of all failures, with omissions far more frequent than insertions.

\begin{table}[t]
\centering
\begin{tabularx}{\columnwidth}{>{\raggedright\arraybackslash}X
                               >{\centering\arraybackslash}X
                               >{\centering\arraybackslash}X}
\toprule
\textbf{Error Type} & \textbf{SIGMORPHON 2020} & \textbf{SIGMORPHON 2023} \\
\midrule
Gemination omission         & 33 (62.3\%) & 32 (62.7\%) \\
Gemination insertion        & 7 (13.2\%)  & 9 (17.6\%)  \\
Phonological substitution   & 6 (11.3\%)  & 4 (7.8\%)   \\
Morpheme boundary           & 3 (5.7\%)   & 4 (7.8\%)   \\
Character recognition (UNK) & 1 (1.9\%)   & 1 (2.0\%)   \\
Compound verb error         & 3 (5.7\%)   & 1 (2.0\%)   \\
\midrule
\textbf{Total}              & \textbf{53} & \textbf{51} \\
\bottomrule
\end{tabularx}
\caption{Quantitative distribution of error types across models and datasets.}
\label{tab:error-quant}
\end{table}

\subsection{Verb-Class Asymmetry}

Error rates vary markedly by verb class. As shown in Table~\ref{tab:error-verbclass-acl}, Type 4‑2 verbs are massively overrepresented in residual errors (30–43\% of errors despite comprising <1\% of the dataset), while Type 4‑1 also exhibits elevated error rates relative to its prevalence. By contrast, Type 1 and Type 2 verbs show error rates roughly proportional to their dataset frequencies. This asymmetric distribution underscores that gemination conditioned by stem‑final /e/ vowels represents a particular structural vulnerability for neural inflection models.

\begin{table}[t]
\centering
\begin{tabularx}{\columnwidth}{>{\centering\arraybackslash}X
                               >{\centering\arraybackslash}X
                               >{\centering\arraybackslash}X
                               >{\centering\arraybackslash}X}
\toprule
\textbf{Verb Type} & \textbf{2020 Errors} & \textbf{2023 Errors} & \textbf{Dataset Count} \\
\midrule
Type 1   & 15 (28.3\%) & 13 (25.5\%) & 2,503 \\
Type 2   & 9 (17.0\%)  & 10 (19.6\%) & 1,298 \\
Type 4-1 & 13 (24.5\%) & 6 (11.8\%)  & 119   \\
Type 4-2 & 16 (30.2\%) & 22 (43.1\%) & 37    \\
Type 4-3 & 0           & 0           & 1     \\
\midrule
\textbf{Total} & \textbf{53} & \textbf{51} & \textbf{3,958} \\
\bottomrule
\end{tabularx}
\caption{Error distribution by verb type.}
\label{tab:error-verbclass-acl}
\end{table}

\subsection{Cross-Model Consistency}

Despite architectural differences between the models (standard shared-task baseline for 2020 and lemma-split training regime for 2023), the observed error patterns are highly consistent:

\begin{itemize}
    \item 75–80\% gemination-related failures
    \item Comparable rates of phonological substitution and morpheme boundary errors
    \item Overlapping problematic verb sets, notably Type 4-2
    \item Identical character recognition failures
    \item High cross-seed stability
\end{itemize}

This convergence suggests that the failures reflect systematic properties associated with orthographic representation in Japanese past-tense formation in \textit{hiragana}, rather than being specific to a single model architecture.

\subsection{Qualitative Error Patterns}

Residual errors in \textit{hiragana}-based inflection exhibit clear, linguistically interpretable patterns. Gemination omission is the most frequent failure, particularly in Type 4-2 verbs, where the small \langjp{っ} is not inserted before the stem-final /e/. For example, \langjp{あきれかえった} \textit{akirekaetta} → \langjp{あきれかえた} \textit{akirekaeta} illustrates this omission. Gemination insertion errors occur mainly in Type 2 verbs, exemplified by \langjp{おきた} \textit{okita} → \langjp{おきった} \textit{okitta}, where an unnecessary \langjp{っ} is inserted.  

Stem alternation and overregularization errors reflect incomplete modeling of consonant deletion or overgeneralization of regular patterns. Examples include \langjp{まった} \textit{matta} → \langjp{まつった} \textit{matsutta}, highlighting the model’s difficulty in correctly applying context-sensitive stem modifications.

Morpheme boundary errors arise when compound boundaries are unmarked in \textit{hiragana}, creating decoding ambiguity; for example, \langjp{ほめたたえた} \textit{hometataeta} → \langjp{ほめたえた} \textit{hometaeta}.  

Character recognition failures are rare and typically involve low-frequency characters. For instance, a single UNK error appeared in both models: \langjp{つっぷした} → \langjp{つっ<UNK>した}, showing that extremely rare symbols can still cause failures even in otherwise high-performing models.  

These qualitative patterns, together with the quantitative distributions reported in Section 4.3, demonstrate that errors are systematically tied to orthographic and morphophonological properties of \textit{hiragana} rather than random model noise. By linking error types to specific verb classes and orthographic phenomena, this analysis highlights structured vulnerabilities that persist even in high-performing character-level neural models.

\section{Discussion}

Residual errors in past-tense inflection reveal systematic vulnerabilities associated with \textit{hiragana} orthography. In particular, verbs whose stems end in /e/ and require gemination before the past-tense suffix account for a disproportionate share of errors relative to their dataset frequency, suggesting that these failures reflect orthographic and morphophonological complexity rather than random variation.

Several factors contribute to these structured errors. First, consonant doubling (small \textit{\langjp{っ}}) introduces an additional mora explicitly represented in \textit{hiragana}, altering the word’s rhythmic structure. Character-level models must generate this symbol in the correct context, a process that interacts with stem alternations and suffix placement in ways that reflect underlying linguistic structure \citep{PimentelEtAl2020}. Second, the graphemic form alone does not encode all conditioning factors, such as lexical or frequency patterns, making some verbs more prone to errors. Third, rare patterns and subtle orthographic contrasts can be dominated by frequent forms, producing persistent error clusters not apparent from aggregate accuracy alone \citep{BelinkovBisk2018}.

Across multiple random seeds and two model architectures, error distributions remain consistent: the same verbs and orthographic processes fail repeatedly. This stability suggests that orthographic devices such as gemination are consistently associated with difficulties in character-level modeling. These findings suggest the potential value of inductive biases that explicitly encode moraic structure.

Overall, our findings highlight the importance of orthography-aware evaluation. Without it, structured error patterns tied to specific orthographic phenomena, such as consonant gemination, may be obscured by high aggregate accuracy, limiting our understanding of neural models in morphologically complex languages.

\section{Limitations}

Several limitations of this study warrant discussion. First, the analysis focuses exclusively on the past-tense paradigm; other inflectional forms, such as negative, passive, or causative, may exhibit different error profiles. Second, our models operate solely on \textit{hiragana} representations. While this reduces orthographic variability, interactions with \textit{kanji} or mixed-script usage could produce different model behavior. Finally, our analysis does not fully disentangle orthographic representation effects from data sparsity and frequency-driven generalization.

We therefore interpret our findings as reflecting correlational patterns among orthographic representation, morphological structure, and frequency effects. Controlled comparisons across alternative input representations (e.g., IPA or romanization) are left for future work.

\section{Conclusion}

We provide an orthography-aware analysis of Japanese past-tense inflection in \textit{hiragana}. While models achieve high overall accuracy, residual errors are structured and linguistically interpretable, with a strong concentration on orthographic devices such as consonant gemination. These errors are consistent with interactions between orthographic representation, morphophonological structure, and model learning dynamics, rather than being attributable to random variation. Their stability across seeds and model architectures suggests robust and systematic error patterns associated with these factors.

Our findings highlight that aggregate accuracy alone can mask systematic orthography-sensitive error patterns. Evaluating performance through the lens of orthographic representation and morphological structure reveals challenges that neural models face in generating morphologically complex forms, and may inform the development of more linguistically informed approaches to morphological generation.

\section{Acknowledgments}

We thank the reviewers and colleagues for their feedback.

\section{References}
\label{sec:reference}

\nocite{zhang2026irregularity}

\bibliographystyle{lrec2026-natbib}
\bibliography{lrec2026-example}

@InProceedings{VaswaniEtAl2017,
  author    = {Vaswani, Ashish and Shazeer, Noam and Parmar, Niki and Uszkoreit, Jakob and Jones, Llion and Gomez, Aidan N. and Kaiser, {\L}ukasz and Polosukhin, Illia},
  title     = {Attention is all you need},
  booktitle = {Advances in Neural Information Processing Systems 30 (NeurIPS 2017)},
  pages     = {5998--6008},
  publisher = {Curran Associates, Inc.},
  year      = {2017}
}

@InProceedings{WuCotterell2019,
  author    = {Wu, Shijie and Cotterell, Ryan},
  title     = {Exact hard monotonic attention for character-level transduction},
  booktitle = {Proceedings of the 57th Annual Meeting of the Association for Computational Linguistics},
  pages     = {1530--1537},
  address   = {Florence, Italy},
  publisher = {Association for Computational Linguistics},
  year      = {2019}
}

@InProceedings{GoldmanEtAl2023,
  author    = {Goldman, Omer and Batsuren, Khuyagbaatar and Khalifa, Salam and Arora, Aryaman and Nicolai, Garrett and Tsarfaty, Reut and Vylomova, Ekaterina},
  title     = {{SIGMORPHON}–{UniMorph} 2023 shared task 0: Typologically diverse morphological inflection},
  booktitle = {Proceedings of the 20th SIGMORPHON Workshop on Computational Research in Phonetics, Phonology, and Morphology},
  pages     = {117--125},
  address   = {Toronto, Canada},
  publisher = {Association for Computational Linguistics},
  year      = {2023}
}

@InProceedings{VylomovaEtAl2020,
  author    = {Vylomova, Ekaterina and White, Jennifer and Salesky, Elizabeth and Mielke, Sabrina J. and Wu, Shijie and Ponti, Edoardo Maria and Maudslay, Rowan Hall and Zmigrod, Ran and Valvoda, Josef and Toldova, Svetlana and Tyers, Francis and Klyachko, Elena and Yegorov, Ilya and Krizhanovsky, Natalia and Czarnowska, Paula and Nikkarinen, Irene and Krizhanovsky, Andrew and Pimentel, Tiago and Hennigen, Lucas Torroba and Kirov, Christo and Nicolai, Garrett and Williams, Adina and Anastasopoulos, Antonios and Cruz, Hilaria and Chodroff, Eleanor and Cotterell, Ryan and Silfverberg, Miikka and Hulden, Mans},
  title     = {{SIGMORPHON} 2020 shared task 0: Typologically diverse morphological inflection},
  booktitle = {Proceedings of the 17th SIGMORPHON Workshop on Computational Research in Phonetics, Phonology, and Morphology},
  pages     = {1--39},
  address   = {Online},
  publisher = {Association for Computational Linguistics},
  year      = {2020}
}

@InProceedings{CotterellEtAl2016,
  author    = {Cotterell, Ryan and Kirov, Christo and Sylak-Glassman, John and Yarowsky, David and Eisner, Jason and Hulden, Mans},
  title     = {The {SIGMORPHON} 2016 shared task—morphological reinflection},
  booktitle = {Proceedings of the 14th SIGMORPHON Workshop on Computational Research in Phonetics, Phonology, and Morphology},
  pages     = {10--22},
  publisher = {Association for Computational Linguistics},
  year      = {2016}
}

@InProceedings{PimentelEtAl2020,
  author    = {Pimentel, Tiago and Valvoda, Josef and Maudslay, Rowan Hall and Zmigrod, Ran and Williams, Adina and Cotterell, Ryan},
  title     = {Information-theoretic probing for Linguistic structure},
  booktitle = {Proceedings of the 58th Annual Meeting of the Association for Computational Linguistics},
  pages     = {4609--4622},
  address   = {Online},
  publisher = {Association for Computational Linguistics},
  year      = {2020}
}

@InProceedings{MakarovClematide2018,
  author    = {Makarov, Peter and Clematide, Simon},
  title     = {Imitation learning for neural morphological string transduction},
  booktitle = {Proceedings of the 2018 Conference on Empirical Methods in Natural Language Processing},
  pages     = {2877--2882},
  address   = {Brussels, Belgium},
  publisher = {Association for Computational Linguistics},
  year      = {2018}
}

@InProceedings{BelinkovBisk2018,
  author    = {Belinkov, Yonatan and Bisk, Yonatan},
  title     = {Synthetic and Natural Noise both break neural machine translation},
  booktitle = {Proceedings of the 6th International Conference on Learning Representations (ICLR 2018)},
  year      = {2018}
}

@Book{Sproat2000,
  author    = {Sproat, Richard},
  title     = {A Computational Theory of Writing Systems},
  publisher = {Cambridge University Press},
  year      = {2000}
}

@Book{DanielsBright1996,
  author    = {Daniels, Peter T. and Bright, William},
  title     = {The World’s Writing Systems},
  publisher = {Oxford University Press},
  address   = {Oxford},
  year      = {1996}
}

@Book{Vance1989,
  author    = {Vance, Timothy J.},
  title     = {An Introduction to {J}apanese Phonology},
  publisher = {State University of New York Press},
  year      = {1989}
}

@Book{Labrune2012,
  author    = {Labrune, Laurence},
  title     = {The Phonology of Japanese},
  publisher = {Oxford University Press},
  year      = {2012}
}

@InProceedings{KubozonoEtAl2008,
  author    = {Kubozono, Haruo and Ito, Junko and Mester, Armin},
  title     = {Consonant gemination in {J}apanese loanword phonology},
  booktitle = {Proceedings of the 18th International Congress of Linguistics (Seoul)},
  pages     = {953--973},
  year      = {2008}
}

@InProceedings{Zhang-2023,
  author    = {Zhang, Wen},
  title     = {Pronunciation Ambiguities in {J}apanese {K}anji},
  booktitle = {Proceedings of the Workshop on Computation and Written Language (CAWL 2023)},
  year      = {2023},
  pages     = {50--60},
  address   = {Toronto, Canada},
  publisher = {Association for Computational Linguistics}
}

@InProceedings{ReimersGurevych2017,
  author    = {Reimers, Nils and Gurevych, Iryna},
  title     = {Reporting score distributions makes a difference: Performance study of {LSTM}-networks for sequence tagging},
  booktitle = {Proceedings of the 2017 Conference on Empirical Methods in Natural Language Processing},
  pages     = {338--348},
  address   = {Copenhagen, Denmark},
  publisher = {Association for Computational Linguistics},
  year      = {2017}
}

@misc{zhang2026irregularity,
  author       = {Zhang, Wen},
  title        = {When Irregularity Helps: A Subclass Analysis of Inductive Bias in Neural Morphology},
  howpublished = {\textit{arXiv}},
  year         = {2026},
  note         = {\href{https://arxiv.org/abs/2605.20558}{arXiv:2605.20558}}
}
\end{document}